\documentclass[10pt,journal,compsoc]{IEEEtran}

\ifCLASSOPTIONcompsoc
  \usepackage[nocompress]{cite}
\else
  \usepackage{cite}
\fi

\usepackage[pdftex]{graphicx}
\usepackage{amsmath,amsfonts,amssymb,bm}
\usepackage{amsthm}
\usepackage{stmaryrd}
\usepackage{textcomp}
\usepackage{multirow}
\usepackage{booktabs}
\usepackage{adjustbox}
\usepackage{subcaption}
\usepackage{colortbl}
\usepackage{xspace}
\usepackage[dvipsnames]{xcolor}
\usepackage[pagebackref=true,breaklinks=true,bookmarks=false]{hyperref}

\definecolor{Gray}{gray}{0.85}
\def\eg{\textit{e.g.}\xspace}
\def\ie{\textit{i.e.}\xspace}
\newcommand{\ubold}[1]{\fontseries{b}\selectfont#1}
\newcommand{\uda}{\,$\triangleright$}
\newcommand{\x}{\bm{x}}

\newcommand{\y}{\bm{y}}
\newcommand{\bw}{\mathbf{w}}
\newcommand{\btheta}{\bm{\theta}}
\newcommand{\bphi}{\bm{\varphi}}
\newcommand{\bpsi}{\bm{\psi}}
\newcommand{\so}{\text{s}}
\newcommand{\tg}{\text{t}}
\newcommand{\cD}{\mathcal{D}}
\newcommand{\cX}{\mathcal{X}}
\newcommand{\cY}{\mathcal{Y}}
\newcommand{\cL}{\mathcal{L}}
\newcommand{\xso}{\x_{\so}}
\newcommand{\xson}{\x_{\so,n}}
\newcommand{\xtg}{\x_{\tg}}
\newcommand{\xtgn}{\x_{\tg,n}}
\newcommand{\Pmap}{\mathsf{P}_{\x}^{\bw}}

\newcommand{\Pmapt}{\mathsf{P}_{\xtg}^{\bw}}
\newcommand{\Cmap}{\mathsf{C}_{\x}^{\btheta}}
\newcommand{\Cmapt}{\mathsf{C}_{\xtg}^{\btheta}}

\newcommand{\Cmaps}{\mathsf{C}_{\xso}^{\btheta}}
\newcommand{\Cmapsn}{\mathsf{C}_{\xson}^{\btheta}}

\begin{document}
%
\title{Confidence Estimation via Auxiliary Models}
%
%
%
%

\author{Charles~Corbi\`ere,
        Nicolas~Thome,
        Antoine~Saporta,
        Tuan-Hung~Vu,
        Matthieu~Cord,
        and~Patrick~P\'erez}

\markboth{IEEE TRANSACTIONS ON PATTERN ANALYSIS AND MACHINE INTELLIGENCE}%
{Corbi\`ere \MakeLowercase{\textit{et al.}}: Confidence Learning via Auxiliary Networks}
%

\IEEEtitleabstractindextext{%
\begin{abstract}
Reliably quantifying the confidence of deep neural classifiers is a challenging yet fundamental requirement for deploying such models in safety-critical applications. In this paper, we introduce a novel target criterion for model confidence, namely the true class probability (TCP). We show that TCP offers better properties for confidence estimation than standard maximum class probability (MCP). Since the true class is by essence unknown at test time, we propose to learn TCP criterion from data with an auxiliary model, introducing a specific learning scheme adapted to this context. We evaluate our approach on the task of failure prediction and of self-training with pseudo-labels for domain adaptation, which both necessitate effective confidence estimates. Extensive experiments are conducted for validating the relevance of the proposed approach in each task. We study various network architectures and experiment with small and large datasets for image classification and semantic segmentation. In every tested benchmark, our approach outperforms strong baselines. 
\end{abstract}

}

\maketitle

\IEEEdisplaynontitleabstractindextext

%
\IEEEpeerreviewmaketitle

\IEEEraisesectionheading{\section{Introduction}}
\label{sec:introduction}

\IEEEPARstart{L}{ast} decade's research in deep learning lead to tremendous boosts in predictive performance for various tasks including image classification \cite{krizhevsky2012imagenet}, object recognition \cite{NIPS2015_5638, 44872, DBLP:conf/cvpr/RedmonDGF16}, natural language processing \cite{journals/corr/abs-1301-3781, conf/interspeech/MikolovKBCK10} and speech recognition \cite{hinton2012deep, hannun2014speech}. However, safety remains a great concern when it comes to implementing these models in real-world conditions \cite{DBLP:journals/corr/AmodeiOSCSM16, journals/corr/JanaiGBG17}. Failing to detect possible errors or over-estimating the confidence of a prediction may carry serious repercussions in critical visual-recognition applications such as in autonomous driving, medical diagnosis \cite{medicaldiag2018} or nuclear power plant monitoring \cite{Linda:2009:NNB:1704175.1704190}. 

Classification with a reject option \cite{Chow1957AnOC,bartlettreject2008,NIPS2016_6336}, also known as \emph{selective classification} \cite{elyaniv10a,NIPS2017_7073}, consists in a scenario where the classifier is given the option to reject an instance instead of predicting its label. Equipped with a reject option, a classifier could decide to stick to the prediction or, on the contrary, to hand over to a human or a back-up system with, \eg, other sensors, or simply to trigger an alarm. One common approach for tackling the problem is to discriminate with a confidence-based criterion: For an instance $\bm{x}$, along with a prediction $f(\bm{x})$, a scalar value $g(\bm{x})$ that quantifies the confidence of the classifier in its prediction is also provided.

Correctly identifying uncertain predictions thanks to low confidence values $g(\bm{x})$ could be beneficial for classification improvements in active learning \cite{pmlr-v70-gal17a} or for efficient exploration in reinforcement learning \cite{Gal:2016:DBA:3045390.3045502}. On a related matter, one would expect the confidence criterion to correlate successful predictions with high values. Some paradigms, such as self-training with pseudo-labeling \cite{lee-icml2013, Li_2019_CVPR}, consist in picking and labeling the most confident samples before retraining the network accordingly. The performance improves by selecting successful predictions thanks to an accurate confidence criterion. A final perspective, linked to failure prediction \cite{DBLP:conf/ivs/HeckerDG18, hendrycks17baseline, NIPS2018_7798}, is the capacity of models to provide a ranking which enables distinguishing correct from erroneous predictions. In each of the previous tasks, obtaining reliable estimates of the predictive confidence is then of prime importance. 

Confidence estimation has been explored in a wide variety of applications, including computer vision \cite{hendrycks17baseline,kendall2015bayesian}, speech recognition \cite{latticeRNNspeech1, latticeRNNspeech2, confspeech2011}, reinforcement learning \cite{Gal:2016:DBA:3045390.3045502} or machine translation \cite{Blatz:2004}.
A widely used baseline with neural-network classifiers is to take the value of the predicted class' probability, namely the \textit{maximum class probability} (MCP), given by the softmax layer output. Although recent evaluations of MCP with modern deep models reveal reasonable performance~\cite{hendrycks17baseline}, they still suffer from several conceptual drawbacks. In particular, MCP leads by design to high confidence values, even for erroneous predictions, since the largest softmax output is used. This design tends to make erroneous and correct predictions overlap in terms of confidence and thus limits the capacity to distinguish them. 

In this work, we identify a better confidence criterion, the \emph{true class probability} (TCP), for deep neural network classifiers with a reject option. For a sample $\bm{x}$, TCP corresponds to the probability of the model with respect to the true class $y$ of that sample, which naturally reflects a better-behaved model's confidence. We provide simple guarantees of the quality of this criterion regarding confidence estimation. Since the true class is obviously unknown at test time, we propose a novel approach which consists in designing an auxiliary network specifically dedicated to estimate the confidence of a prediction. Given a trained classifier $f$, this auxiliary network learns the TCP criterion from data. In inference, we use its scalar output as the confidence estimate $g(\bm{x})$ associated with the prediction. When applied to failure prediction, we observe significant improvements over strong baselines. Our approach  is  also  adequate  for  self-training  strategies in unsupervised domain adaptation. To meet the challenge of this task in semantic segmentation, we propose an enhanced architecture with structured output and adopt an adversarial learning scheme which  enforces alignment between confidence maps in source and target domains. A thorough analysis of our approach, including relevant variations, ablation studies and qualitative evaluations of confidence estimates, helps to gain insight about its behavior.

\smallskip
In summary, our contributions are as follows:
\begin{itemize}
    \item We define a novel confidence criterion, the \emph{true class probability}, which exhibits an adequate behavior for confidence estimation;
    \item We propose to design an auxiliary neural network, coined \emph{ConfidNet}, which aims to learn this confidence criterion from data;
    \item We apply this approach to the task of failure prediction and to self-training in unsupervised domain adaptation with adequate choices of architecture, loss function and learning scheme;
    \item We extensively experiment across various benchmarks and backbone networks to validate the relevance of our approach on both tasks.
\end{itemize}

The paper is organized as follows. In Section~\ref{sec:related_work}, we provide an  overview  of  the  most  relevant  related  works on confidence estimation, failure prediction, self-training and unsupervised domain adaptation. Section~\ref{sec:learning_confidence} exposes our approach for confidence estimation based on learning an adequate criterion via an auxiliary network. We also describe how it relates to classification with a reject option. In Section~\ref{sec:confidnet}, we adapt our approach to failure prediction by introducing an architecture, a loss function and a learning scheme for this task. Similarly, Section~\ref{sec:conda} details the instantiation of our approach for confidence-based self-training in unsupervised domain adaptation (DA), which we denote as ConDA. In particular, we present two additions, an adversarial loss and a multi-scale confidence architecture, which further help to improve the performance for this task. Finally, we report experimental studies in Section~\ref{sec:experiments}. This paper extends a previous conference publication \cite{corbiere2019ConfidNet} by introducing: (1) An comprehensive adaptation of the approach to improve the key step of self-training from pseudo-labels in semantic segmentation with DA; (2) An exploration of the classification-with-rejection framework, which strengthens the rationale of the proposed approach.

\section{Related work}
\label{sec:related_work}

\subsection{Confidence estimation}

Confidence estimation in machine learning has been around for many decades, firstly linked to the idea of classification with a reject option \cite{Chow1957AnOC}. Following works \cite{bartlettreject2008,NIPS2016_6336, cortes2016, zaragoza1998confidence} explored alternative rejection criteria. In particular, \cite{cortes2016} proposes to jointly learn the classifier and the selection function. El-Yaniv \cite{elyaniv10a} provides an analysis of the risk-coverage trade-off that occurs when classifying with a reject option. More recently, \cite{NIPS2017_7073, Geifman2019SelectiveNetAD} extend the approach to deep neural networks, considering various confidence measures.

Since the wide adoption of deep learning methods, confidence estimation has raised even more interest as recent works reveal that modern neural networks tend to be overconfident \cite{conf/cvpr/NguyenYC15}, non-calibrated~\cite{GuoPSW17,Neumann18c}, sensitive to adversarial attacks~\cite{goodfellow2014explaining, DBLP:journals/corr/SzegedyZSBEGF13} and inadequate to distinguish in- from out-of-distribution examples~\cite{hendrycks17baseline, liang2018enhancing, NIPS2017_7219}.

Bayesian neural networks \cite{bnn1996} offer a principled approach for confidence estimation by adopting a Bayesian formalism which models the weight posterior distribution. As the true posterior cannot be evaluated analytically in complex models, various approximations have been developed, such as variational inference \cite{Blundell:2015:WUN:3045118.3045290, NIPS2019_9472, Gal:2016:DBA:3045390.3045502} or expectation propagation \cite{pmlr-v37-hernandez-lobatoc15}. In particular, MC Dropout \cite{Gal:2016:DBA:3045390.3045502} has raised a lot of interest due to the simplicity of its implementation. Predictions are obtained by averaging softmax vectors from multiple feed-forward passes through the network with dropout layers. When applied to regression, the predictive distribution uncertainty can be summarized by computing statistics, \eg, variance. However, when using MC Dropout for uncertainty estimation in classification tasks, the predictive distribution is averaged to a point-wise softmax estimate before computing standard uncertainty criteria such as entropy. It is worth mentioning that these entropy-based criteria measure the softmax output dispersion, where the uniform distribution has maximum entropy. It is not clear how well these dispersion measures are adapted to distinguishing failures from correct predictions, especially with deep neural networks which output overconfident predictions~\cite{GuoPSW17}: for example, it might be very challenging to discriminate a peaky prediction corresponding to a correct prediction from an incorrect overconfident one. Lakshminarayanan \textit{et al}. \cite{NIPS2017_7219} propose an alternative to Bayesian neural networks by leveraging an ensemble of neural networks to produce well-calibrated uncertainty estimates. However, it requires training multiple classifiers, which has a considerable computing cost in training and inference time.

\begin{figure*}[t]
\centering
\begin{minipage}[c]{0.45\linewidth}
\centering
    \includegraphics[width=\linewidth]{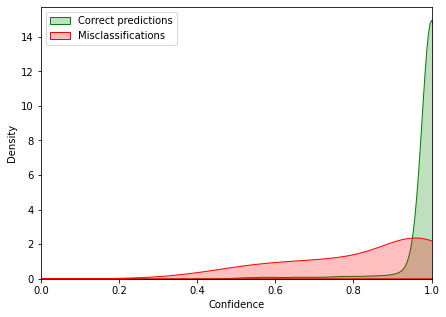}
    \subcaption{Maximum Class Probability}
    \label{fig:density-plot-mcp}
\end{minipage}%
\hspace{0.3cm}
\begin{minipage}{0.45\linewidth}
\centering
    \includegraphics[width=\linewidth]{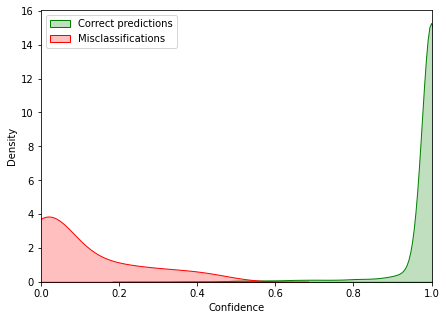}
    \subcaption{True Class Probability}
    \label{fig:density-plot-tcp}
\end{minipage}
\caption{\textbf{Distributions of different confidence measures over correct and erroneous predictions of a given model.} When ranking according to MCP (a) the test predictions of a convolutional model trained on CIFAR-10, we observe that correct ones (in green) and misclassifications (in red) overlap considerably, making it difficult to distinguish them. On the other hand, ranking samples according to TCP (b) alleviates this issue and allows a much better separation.}
\label{fig:density-plot}
\end{figure*}

\subsection{Failure prediction}

In the context of classification, a widely used baseline for failure prediction is to take the value of the predicted class' probability given by the softmax layer output, namely the \emph{maximum class probability} (MCP), suggested by \cite{oldconfmcp1995} and revised by \cite{hendrycks17baseline}. As stated before, MCP presents several limits regarding both failure prediction and out-of-distribution detection, as it outputs unduly high confidence values.

Blatz \textit{et al}. \cite{Blatz:2004} introduce a method for confidence estimation in machine translation by solving a binary classification between correct and erroneous predictions. More recently, Jiang \textit{et al.} \cite{NIPS2018_7798} proposed a new confidence measure, `Trust Score', which measures the agreement between the classifier and a modified nearest-neighbor classifier on the test examples. More precisely, the confidence criterion used in Trust Score is the ratio between the distance from the sample to the nearest class different from the predicted class and the distance to the predicted class. One clear drawback of this approach is its lack of scalability, since computing nearest neighbors in large datasets is extremely costly in both computations and memory. Another more fundamental limitation related to the Trust Score itself is that local distance computation becomes less meaningful in high dimensional spaces~\cite{distance-curse}, which is likely to negatively affect the performances of this method as shown in Section~\ref{subsec:exp_confidnet}. 

In tasks closely related to failure prediction, Guo \textit{et al}. \cite{GuoPSW17}, for confidence calibration, and Liang \textit{et al}. \cite{liang2018enhancing}, for out-of-distribution detection, proposed to use temperature scaling to mitigate confidence values. However, this does not affect the ranking of the confidence score and therefore the separability between errors and correct predictions. DeVries \textit{et al}. \cite{devries2018learning} share with us the same purpose of learning confidence in neural networks. Their work differs from ours by focusing on out-of-distribution detection and learning jointly a distribution confidence score and classification probabilities. In addition, their criterion is based on an interpolation between output probabilities and target distribution whereas we specifically define a criterion suited to failure prediction.

\subsection{Self-training in domain adaptation}

\textbf{Unsupervised Domain Adaptation (UDA).} UDA has received a lot of attention over the past few years because of its importance for a variety of real-world problems, such as robotics or autonomous driving. Most works in this line of research aim at minimizing the discrepancy between the data distributions in source and target domains.

Adopting an adversarial training approach \cite{gradreversal2016} has yielded most recent progress in the semantic segmentation task by producing indistinguishable source-target distributions in the space of features extracted by modern convolutional deep neural nets. To cite a few methods: CyCADA~\cite{Hoffman_cycada2017} first stylizes the source-domain images as target-domain images before aligning source and target in the feature space; AdaptSegNet~\cite{Tsai_adaptseg_2018} constructs a multi-level adversarial network to perform output-space domain adaptation at different feature levels; AdvEnt~\cite{vu2018advent} aligns the entropy of the pixel-wise predictions with an adversarial loss; BDL~\cite{Li_2019_CVPR} learns alternatively an image translation model and  a segmentation model that promote each other. 

\smallskip\noindent \textbf{Self-Training.} Semi-supervised learning designates the general problem where a decision rule must be learned from both labeled and unlabeled data. Among the methods applied to address this problem, self-training with pseudo-labeling~\cite{lee-icml2013} is a simple strategy that relies on picking up the current predictions on the unlabeled data and using them as if they were true labels for further training. It is shown in \cite{lee-icml2013} that the effect of pseudo-labeling is equivalent to entropy regularization~\cite{grandvalet-nips2005}. In a UDA setting, the idea is to collect pseudo-labels on the unlabeled target-domain samples in order to have an additional supervision loss in the target domain. To select only reliable pseudo-labels, such that the performance of the adapted semantic segmentation network effectively improves, BDL~\cite{Li_2019_CVPR} resorts to standard selection with MCP. ESL~\cite{saporta2020esl} uses instead the entropy of the prediction as confidence criterion for its pseudo-label selection. CBST~\cite{zou2018unsupervised} proposes an iterative self-training procedure where the pseudo-labels are generated based on a loss minimization. In~\cite{zou2018unsupervised}, the authors also propose a way to balance the classes in their pseudo-labels to avoid the dominance of large classes as well as a way to introduce spatial priors. More recently, the CRST framework~\cite{Zou_2019_ICCV} proposes multiple types of confidence regularization to limit the propagation of errors caused by noisy pseudo-labels. 

\begin{figure*}[t]
  \centering
  \includegraphics[width=0.95\linewidth]{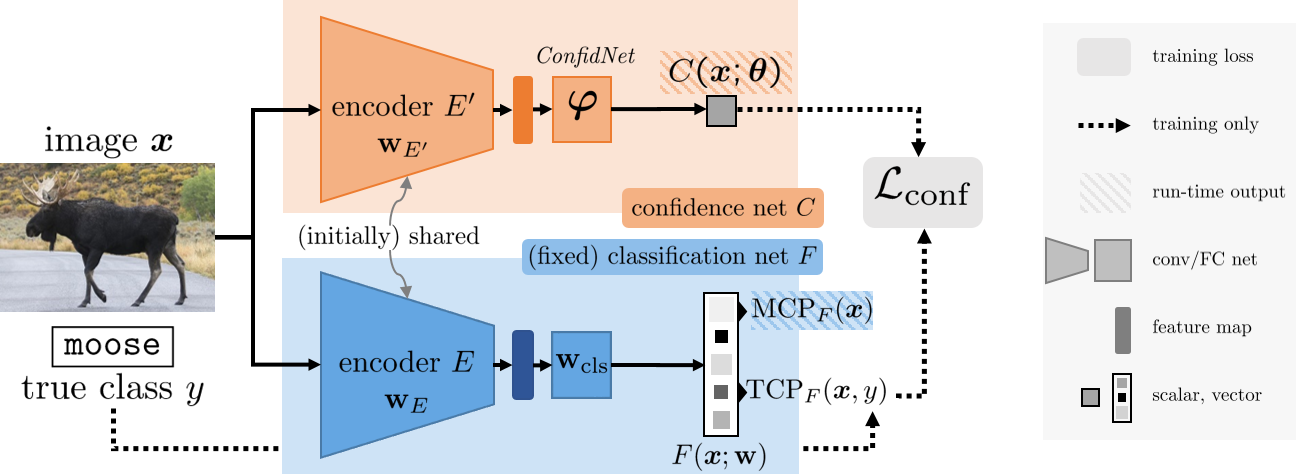}
  \caption{\textbf{Learning confidence approach.} The fixed classification network $F$, with parameters $\bw =(\bw_{E},\bw_{\text{cls}})$, is composed of a succession of convolutional and fully-connected layers (encoder $E$) followed by last classification layers with softmax activation. The auxiliary confidence network $C$, with parameters $\btheta$, builds upon the feature maps extracted by the encoder $E$, or its fine-tuned version $E'$ with parameters $\bw_{\text{E'}}$: they are passed to ConfidNet, a trainable multi-layer module with parameters $\bphi$. The auxiliary model outputs a confidence score  $C(\bm{x};\btheta)\in[0,1]$, with $\btheta = \bphi$ in absence of encoder fine-tuning and $\btheta =(\bw_{E'},\bphi)$ in case of fine-tuning.}
  \label{confidnet_network}
\end{figure*}

\section{Learning a model's confidence with an auxiliary model}
\label{sec:learning_confidence}

In this section, we first briefly introduce the task of classification with a reject option, along with necessary notations.  
We then introduce an effective confidence-rate function for neural-net classifiers and we present our approach to learn this target confidence-rate function thanks to an auxiliary neural network. 
For sake of simplicity, we consider in this section a generic classification task, where the input is raw or transformed signals and the expected output is a predicted category. The semantic segmentation task we address in Section \ref{sec:conda} is in effect a pixel-wise classification of localized features derived from the input image.

\subsection{Problem formulation}
\label{sec:problem_formulation}

 Let us consider a dataset $\cD= \{ (\x_n, y_n) \}_{n=1}^N$ composed of $N$ \textit{i.i.d.} training samples, where $\x_n \in \cX \subset \mathbb{R}^D$ is a $D$-dimensional data representation, deep feature maps from an image or the image itself for instance, and $y_n \in \cY=\llbracket 1, K \rrbracket$ is its true class among the $K$ pre-defined categories. These samples are drawn from an unknown joint distribution $P(X,Y)$ over $(\cX, \cY)$.

A \textit{selective classifier} \cite{elyaniv10a, NIPS2017_7073} is a pair $(f,g)$ where $f: \cX \rightarrow \cY$ is a \textit{prediction function} and $g: \cX \rightarrow \{0,1\}$ is a \textit{selection function} which enables to reject a prediction:
\begin{equation}
    (f,g)(\bm{x}) =
    \begin{cases}
      f(\x), & \text{if}\ g(\x)=1 \, , \\
      \text{reject}, &\text{if}\ g(\x)=0 \, . \\
    \end{cases}
\end{equation}

In this work, we focus on classifiers based on artificial neural networks. Given an input $\x$, such a network $F$ with parameters $\bw$ outputs non-negative scores over all classes, which are normalized through softmax. If well trained, this output can be interpreted as the predictive distribution
$P(Y \vert \x, \hat{\bw}) = F(\x;\hat{\bw}) \in \Delta$, with $\Delta$ the probability $K$-simplex in $\mathbb{R}^{K}$ and $\hat{\bw}$ the learned weights.
Based on this distribution, the predicted sample class is usually the maximum \textit{a posteriori} estimate:  
\begin{equation}
f(\bm{x}) = \mathrm{arg}\!\max_{k \in \cY}~P(Y = k \vert \bm{x}, \hat{\bw}) = \mathrm{arg}\!\max_{k \in \cY} F(\x;\hat{\bw})[k].
\label{eq:F2f}
\end{equation}

\indent We are not interested here in trying to improve the accuracy of the already-trained model $F$, but rather to make its future use more reliable by endowing the system with the ability to recognize when the prediction might be wrong.

To this end, a \textit{confidence-rate function} $\kappa_f:\cX \rightarrow \mathbb{R}^{+}$ is associated to $f$ so as to assess the degree of confidence of its predictions, the higher the value the more certain the prediction \cite{elyaniv10a, NIPS2017_7073}. A suitable confidence-rate function should correlate erroneous predictions with low values and successful predictions with high values. 
Finally, given a user-defined threshold $\delta \in \mathbb{R}^+$, the selection function $g$
can be simply derived from the confidence rate:
\begin{equation}
    g(\bm{x})=
    \begin{cases}
      1 & \text{if}\ \kappa_f(\bm{x}) \geq \delta \, , \\
      0 & \text{otherwise.} \\
    \end{cases}
\end{equation}

\subsection{TCP, an effective confidence-rate function}

For a given input $\x$, a standard confidence-rate function for a classifier $F$ is the probability associated to the predicted max-score class, that is the \textit{maximum class probability}: 
\begin{equation}
\text{MCP}_F(\x) = 
\max_{k \in \cY} P(Y=k \vert \x, \hat{\bw}) =  
\max_{k \in \cY} F(\x;\hat{\bw})[k].
\end{equation} 

However, by taking the largest softmax probability as confidence estimate, MCP leads to high confidence values both for correct and erroneous predictions alike, making it hard to distinguish them, as shown in Figure~\ref{fig:density-plot-mcp}. On the other hand, when the model misclassifies an example, the probability associated to the true class $y$ is lower than the maximum one and likely to be low. Based on this simple observation, we propose to consider instead this \emph{true class probability} as a suitable confidence-rate function.
For any admissible input $\x\in\cX$, we assume the \textit{true} class $y(\x)$ is known, which we denote $y$ for simplicity. The TCP confident rate is defined as  
\begin{equation}
    \text{TCP}_F(\x,\,y) = P(Y=y \vert\,\x, \hat{\bw}) = F(\x;\hat{\bw})[y].
\end{equation}

\smallskip\noindent \textbf{Simple guarantees.} With TCP, the following properties hold (see derivation in Appendix A.1). Given a properly labelled example $(\bm{x},y)$, then:
\begin{itemize}
    \item $\text{TCP}_F(\bm{x},y)> 1/2$ $\Rightarrow$ $f(\x) = y$, \ie the example is correctly classified by the model;
    \item $\text{TCP}_F(\bm{x},y) < 1/K$ $\Rightarrow$ $f(\x) \neq y$, \ie the example is wrongly classified by the model,
\end{itemize} 
where class prediction $f(\x)$ is defined by (\ref{eq:F2f}).

Within the range $[1/K, 1/2]$, there is no guarantee that correct and incorrect predictions will not overlap in terms of TCP. However, when using deep neural networks, we observe that the actual overlap area is extremely small in practice, as illustrated in Figure~\ref{fig:density-plot-tcp} on the CIFAR-10 dataset. One possible explanation comes from the fact that modern deep neural networks output overconfident predictions and therefore non-calibrated probabilities~\cite{GuoPSW17}. We provide consolidated results and analyses on this aspect in Section~\ref{sec:experiments} and in Appendix A.2.

\subsection{Learning to predict TCP with a neural network}

Using TCP as a confidence-rate function on a model's output would be of great help when it comes to reliably estimate its confidence. However, the true classes $y$ are obviously not available when estimating confidence on test inputs. 

We propose to \emph{learn TCP confidence from data}. More formally, for the classification task at hand, we consider a parametric selective classifier $(f,g)$, with $f$ based on an already-trained neural network $F$. We aim at deriving its companion selection function $g$ from a learned estimate of the TCP function of $F$. 
To this end, we introduce an \textit{auxiliary model} $C$, with parameters $\btheta$, that is intended to predict $\text{TCP}_F$ and to act as a confidence-rate function for the selection function $g$. 
An overview of the proposed approach is available in Figure~\ref{confidnet_network}. This model is trained such that, at runtime, for an input $\x\in\cX$ with (unknown) true label $y$, we have: 
\begin{equation}
    C(\x;\btheta) \approx \text{TCP}_F(\x,y).
\end{equation}

\indent In practice, this auxiliary model $C$ will be a neural network trained under full supervision on $\cD$ to produce this confidence estimate. To design this network, we can transfer knowledge from the already-trained classification network. Throughout its training, $F$ has indeed learned to extract increasingly-complex features that are fed to its final classification layers. Calling $E$ the encoder part of $F$, a simple way to transfer knowledge consists in defining and training a multi-layer head with parameters $\bphi$ that regresses $\mathrm{TCP}_F$ from features encoded by $E$. We call \textit{ConfidNet} this module. As a result of this design, the complete confidence network $C$ is composed of a frozen encoder followed by trained ConfidNet layers. As we shall see in Section \ref{sec:confidnet}, the complete architecture might be later fine-tuned, including the encoder, as in classic transfer learning. In that case, $\btheta$ will encompass the parameters of both the encoder and the ConfidNet's layers. 

In the rest of the paper, we detail the different network architectures, loss functions and learning schemes of ConfidNet for two distinct applications: Classification failure prediction and self-training for semantic segmentation with domain adaptation. 
In both tasks, a ranking of unlabelled samples that allows a clear distinction of correct predictions from erroneous ones is crucial.  The proposed auxiliary model offers a new solution to this problem.

\section{Application to failure prediction}
\label{sec:confidnet}

 Given a trained model, failure prediction is the task of predicting at run-time whether the model has taken a correct decision or not for a given input. As discussed in Section \ref{sec:related_work}, there are different ways to attack this task, which has many real-world applications in safety-critical systems especially. With a confidence-rate function in hand, the task can be simply set as thresholding this function, exactly in the same way the selection function works in prediction with a reject option. In this Section, we discuss how ConfidNet can be used for that exact purpose in the context of image classification.

\subsection{Architecture}

State-of-art image classification models are composed of convolutional layers followed by one or more fully-connected layers and a final softmax operation. In order to work with such a classification network $F$, we build ConfidNet upon a late intermediate representation of $F$. ConfidNet is designed as a small multilayer perceptron composed of a succession of dense layers with a final sigmoid activation that outputs $C(\bm{x};\btheta)\in[0,1]$. As explained in Section \ref{sec:learning_confidence}, we will train this network in a supervised manner, such that it predicts well the true-class probability assigned by $F$ to the input image. Regarding the capacity of ConfidNet, we have empirically found that increasing further its depth leaves performance unchanged for estimating the confidence of the classification network (see Appendix B.4 for more details).

\subsection{Loss function}

As we want to regress a score between $0$ and $1$, we use a mean-square-error (MSE) loss to train the confidence model:
\begin{equation} 
\label{eq:loss-conf}
\cL_{\text{conf}}(\btheta;\cD) = \frac{1}{N} \sum_{n=1}^N \big(C(\x_n;\btheta) - \text{TCP}_F(\x_n,y_n)\big)^2.
\end{equation}

Since the final task here is the prediction of failures, with confidence prediction being only a means toward it, a more explicit supervision with failure/success information could be considered. In that case, the previous regression loss could still be used, with 0 (failure) and 1 (success) target values instead of TCP. Alternatively, a binary cross entropy loss (BCE) for the error-prediction task using the predicted confidence as a score could be used. Seeing failure detection as a ranking problem, where good predictions must be ranked before erroneous ones according to the predicted confidence, a batch-wise ranking loss can also be utilized~\cite{Mohapatra_2018_CVPR}. We experimentally assessed all these alternative losses, including a focal version \cite{focaloss} of the BCE to focus on hard examples, as discussed in Section~\ref{subsec:learning_variants}. They lead to inferior performance compared to using (\ref{eq:loss-conf}). This might be due to the fact that TCP conveys more detailed information than a mere binary label on the quality of the classifier's prediction for a sample. Hinton \textit{et al}. \cite{distillation} make a similar observation when using soft targets in knowledge distillation. In situations where only very few error samples are available, this fine-grained information improves the performance of the final failure detection (see Section~\ref{subsec:learning_variants}).

\begin{figure*}[t]
\begin{center}
\includegraphics[width=\linewidth]{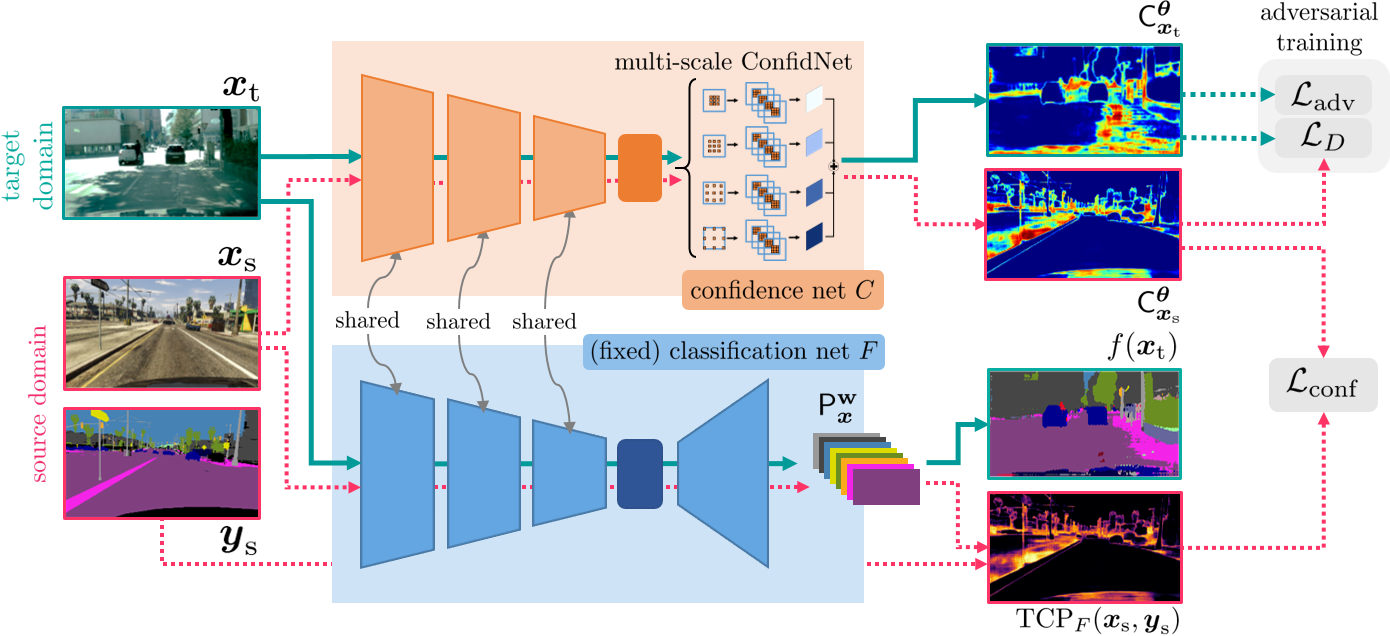}
\end{center}
   \caption{\textbf{Overview of proposed confidence learning for domain adaptation (ConDA) in semantic segmentation}. Given images in source and target domains, we pass them to the encoder part of the segmentation network $F$ to obtain their feature maps. This network $F$ is fixed during this phase and its weights are not updated. The confidence maps are obtained by feeding these feature maps to the trainable head of the confidence network $C$, which includes a multi-scale ConfidNet module. For source-domain images, a regression loss $\cL_{\text{conf}}$ (\ref{eq:loss-conf-conda}) is computed to minimize the distance between $\Cmaps$ and the fixed true-class-probability map $\text{TCP}_F(\xso, \y_{\so})$. An adversarial training scheme -- based on discriminator's loss $\cL_D(\bpsi)$ (Eq.\,\ref{eq:l_Dconf}) and adversarial part $\cL_{\text{adv}}(\btheta)$ of confidence net's loss (Eq.\,\ref{eq:l_C}) --, is also added to enforce the consistency between the $\Cmaps$'s and $\Cmapt$'s. Dashed arrows stand for paths that are used only at train time.
   }
\label{fig:confidence_training}
\end{figure*}

\subsection{Learning scheme}
\label{subsec:confidnet-learning}

We decompose the parameters of the classification network $F$ into $\bw = (\bw_{E}, \bw_{\text{cls}})$, where $\bw_{E}$ denotes its encoder's weights and $\bw_{\text{cls}}$ the weights of its last classification layers. Such as in transfer learning, the training of the confidence network $C$ starts by fixing the shared encoder and training only ConfidNet's  weights $\bphi$. In this phase, the loss (\ref{eq:loss-conf}) is thus minimized only w.r.t. $\btheta = \bphi$.

In a second phase, we further fine-tune the complete network $C$, including its encoder which is now untied from the classification encoder $E$ (the main classification model must remain unchanged, by definition of the addressed problem). Denoting $E'$ this now independent encoder, and $\bw_{E'}$ its weights, this second training phase optimizes (\ref{eq:loss-conf}) w.r.t. $\btheta = (\bw_{E'}, \bphi)$ with $\bw_{E'}$ initially set to $\bw_{E}$.

We also deactivate dropout layers in this last training phase and reduce learning rate to mitigate stochastic effects that may lead the new encoder to deviate too much from the original one used for classification. Data augmentation can thus still be used. ConfidNet can be trained using either the original training set or a validation set. The impact of this choice is evaluated in Section~\ref{subsec:learning_variants}.

In Section \ref{sec:experiments}, we put this framework at work on several standard image-classification benchmarks and analyse its effectiveness in comparison with alternative approaches.

\section{Application to self-training in semantic segmentation with domain adaptation}
\label{sec:conda}

Unsupervised domain adaptation for semantic segmentation aims to adapt a segmentation model trained on a labeled source domain to a target domain devoid of annotation. Formally, let us consider the annotated source-domain training set $\cD_{\so} = \{ (\xson, \y_{\so,n}) \}_{n=1}^{N_{\so}}$, where $\xson$ is a color image of size $(H,W)$ and $\y_{\so,n} \in \cY^{H\times W}$ its associated ground-truth segmentation map. A segmentation network $F$ with parameters $\bw$ takes as input an image $\x$ and returns a predicted \textit{soft}-segmentation map $F(\x;\bw) = \Pmap  \in [0,1]^{H\times W\times K}$, where  $\Pmap[h,w,:] = P(Y[h,w] \,\vert\, \x;\bw)\in\Delta$. 
The final prediction of the network is the segmentation map $f(\x)$ defined pixel-wise as $f(\x)[h,w] = \arg\!\max_{k\in\cY} \Pmap[h,w,k]$.
This network is learned with full supervision from the source-domain samples in $\cD_{\so}$, using a cross-entropy loss, while leveraging a set $\cD_{\tg}$ of unlabelled target-domain examples.

\subsection{Self-training for unsupervised domain adaptation}

In UDA, the main challenge is to use the unlabeled target set $\cD_{\tg} = \{\bm{x}_{\tg,n}\}_{n=1}^{N_{\text{t}}}$ available during training to learn domain-invariant features on which the segmentation model would behave similarly in both domains.
As reviewed in Section \ref{sec:related_work}, a variety of techniques have been proposed to do that, in particular for the task of semantic segmentation. Leveraging automatic pseudo-labeling of target-domain training examples is in particular a simple, yet powerful way to further improve UDA performance with self-training. One key ingredient of such an approach being the selection of the most promising pseudo-labels, the proposed auxiliary confidence-prediction model lends itself particularly well to this task. 
In the rest of this section, we detail how the proposed approach to confidence prediction can be adapted to semantic segmentation, with application to domain adaptation through self-training. The resulting framework, called ConDA, is illustrated in Figure \ref{fig:confidence_training}.   

A high-level view of self-training for semantic segmentation with UDA is a follows:
\begin{enumerate}
    \item Train a segmentation network for the target domain using a chosen UDA technique;\label{step1}
    \item Collect pseudo-labels among the predictions that this network makes on the target-domain training images;\label{step2}
    \item Train a new semantic-segmentation network from scratch using the chosen UDA technique in combination with supervised training on target-domain data with pseudo-labels;
    \item Possibly, repeat from step~\ref{step2} by collecting better pseudo-labels after each iteration.
\end{enumerate}

While the general idea of self-training is simple and intuitive, collecting good pseudo-labels is quite tricky: If too many of them correspond to erroneous predictions of the current segmentation network, the performance of the whole UDA can deteriorate. Thus, a measure of confidence should be used in order to only gather reliable predictions as pseudo-labels and to reject the others.

\subsection{Selecting pseudo-labels with a confidence model}

Following the self-training framework previously described, a confidence network $C$ is learned at step~(\ref{step2}) to predict the confidence of the UDA-trained semantic segmentation network $F$ and used to select only trustworthy pseudo-labels on target-domain images. To this end, the framework proposed in Section \ref{sec:learning_confidence} in an image classification setup, and applied to predicting erroneous image classification in Section \ref{sec:confidnet}, needs here to be adapted to the structured output of semantic segmentation.

Semantic segmentation can be seen as a pixel-wise classification problem. 
Given a target-domain image $\xtg$, we want to predict both its soft semantic map $F(\xtg;\bw)$ and, using an auxiliary model with trainable parameters $\btheta$, its confidence map $C(\xtg;\btheta) = \Cmapt \in [0,1]^{H\times W}$. Given a pixel $(h,w)$, if its confidence $\Cmapt[h,w]$ is above a chosen threshold $\delta$, we label it with its predicted class $f(\xtg)[h,w] = \arg\!\max_{k\in\cY} \Pmapt[h,w,k]$, otherwise it is masked out. Computed over all images in $\cD_{\tg}$, these incomplete segmentation maps constitute target pseudo-labels that are used to train a new semantic-segmentation network. Optionally, we may repeat from step~(\ref{step2}) and learn alternately a confidence model to collect pseudo-labels and a segmentation network using this self-training.

\subsection{Confidence training with adversarial loss}
\label{sec:adv-loss}

To train the segmentation confidence network $C$, we propose to jointly optimize two objectives. Following the approach proposed in Section \ref{sec:learning_confidence}, the first one supervises the confidence prediction on annotated source-domain examples using the known true class probabilities for the predictions from $F$. Specific to semantic segmentation with UDA, the second one is an adversarial loss that aims at reducing the domain gap between source and target. A complete overview of the approach is provided in Figure \ref{fig:confidence_training}.  

\smallskip\noindent\textbf{Confidence loss.} The first objective is a pixel-wise version of the confidence loss in (\ref{eq:loss-conf}). On annotated source-domain images, it requires $C$ to predict at each pixel the score assigned by $F$ to the (known) true class:
\begin{equation} 
\label{eq:loss-conf-conda}
\cL_{\text{conf}}(\btheta;\cD_{\so}) = 
\frac{1}{N_{\so}} \sum_{n=1}^{N_{\so}} 
\big\| 
\Cmapsn - \text{TCP}_F(\xson,\y_{\so,n}) \big\|^2_{\text{F}}, 
\end{equation}
where $\|\cdot\|_{\text{F}}$ denotes the Frobenius norm and, for an image $\x$ with true segmentation map $\y$ and predicted soft one $F(\x;\hat{\bw})$, we note
\begin{equation}
    \text{TCP}_F(\x,\y)[h,w] = F(\x;\hat{\bw})\Big[h,w,\y[h,w]\Big]
\end{equation}
at location $(h,w)$. On a new input image, $C$ should predict at each pixel the score that $F$ will assign to the unknown true class, which will serve as a confidence measure.

However, compared to the application in previous Section, we have here the additional problem of the gap between source and target domains, an issue that might affect the training of the confidence model as in the training of the segmentation model.

\smallskip\noindent \textbf{Adversarial loss.} The second objective concerns the domain gap. While model $C$ learns to estimate TCP on source-domain images, its confidence estimation on target-domain images may suffer dramatically from this domain shift. As classically done in UDA, we propose an adversarial learning of our auxiliary model in order to address this problem. More precisely, we want the confidence maps produced by $C$ in the source domain to resemble those obtained in the target domain.

A discriminator $D:[0,1]^{H \times W} \rightarrow \{0,1\}$, with parameters $\bpsi$, is trained concurrently with $C$ with the aim to recognize the domain (1 for source, 0 for target) of an image given its confidence map. The following loss is minimized w.r.t. $\bpsi$:
\begin{align}
    \cL_D(\bpsi;\cD_{\so}\cup\cD_{\tg}) = 
    &\frac{1}{N_{\so}}\sum\limits_{n=1}^{N_{\so}} \cL_\text{adv}(\xson,1) + \nonumber \\
    &\frac{1}{N_{\tg}}\sum\limits_{n=1}^{N_{\tg}} \cL_\text{adv}(\xtgn,0),
        \label{eq:l_Dconf}
\end{align}
where $\cL_\text{adv}$ denotes the cross-entropy loss of the discriminator based on confidence maps:
\begin{equation}
    \cL_\text{adv}(\x,\lambda) = -\lambda\log\big(D(\Cmap;\bpsi)\big) - (1-\lambda)\log(1-D\big(\Cmap;\bpsi)\big),
\end{equation}
for $\lambda= \{0,1\}$, which is a function of both $\bpsi$ and $\btheta$. In alternation with the training of the discriminator using (\ref{eq:l_Dconf}), the adversarial training of the confidence net is conducted by minimizing, w.r.t. $\btheta$, the following loss:
\begin{equation}
    \cL_C(\btheta;\cD_{\so}\cup\cD_{\tg}) = \cL_\text{conf}(\btheta; \cD_{\so}) + \frac{\lambda_\text{adv}}{N_{\tg}}\sum\limits_{n=1}^{N_{\tg}}\cL_\text{adv}(\xtg,1),
    \label{eq:l_C}
\end{equation}
where the second term, weighted by $\lambda_\text{adv}$, encourages $C$ to produce maps in target domain that will confuse the discriminator. 

This adversarial scheme for confidence learning also acts as a regularizer during training, improving the robustness of the unknown TCP target confidence. As the training of $C$ may actually be unstable, adversarial training provides additional information signal, in particular imposing that confidence estimation should be invariant to domain shifts. We empirically observed that this adversarial confidence learning provides better confidence estimates and improves convergence and stability of the training scheme.

\begin{table*}[t]
  \caption{\textbf{Comparison of confidence estimation methods for failure prediction and selective classification}. For each dataset, all methods share the same classification network. For MC Dropout, test accuracy is averaged through random sampling. The first three metrics are percentages and concern failure prediction. The two last ones (the lower, the better) concern selective classification and their values have been multiplied by $10^3$ for clarity. Scores are averaged over 5 runs, best results are in \textbf{bold}, second best ones are \underline{underlined}.}
  \label{comparative-results}
  \centering
  \begin{adjustbox}{max width=\textwidth}
  \begin{tabular}{cl|rrr|rr}
    \toprule
    Dataset & Model & FPR\,@\,95\%\,TPR\,$\downarrow$& AUPR \,$\uparrow$ & AUROC \,$\uparrow$& AURC \,$\downarrow$& E-AURC\,$\downarrow$ \\
    \midrule
    \multirow{4}{*}{\shortstack[c]{\ubold{MNIST} \\ MLP}} & MCP~\cite{hendrycks17baseline} & 14.88 {\scriptsize $\pm 1.42$} & 47.25 {\scriptsize $\pm 1.67$} & 97.28 {\scriptsize $\pm 0.20$} & 0.83 {\scriptsize $\pm 0.07$} & 0.61 {\scriptsize $\pm 0.06$} \\
    & MC Dropout~\cite{Gal:2016:DBA:3045390.3045502} & 15.17 {\scriptsize $\pm 1.08$} & 40.98 {\scriptsize $\pm 1.24$} & 97.10 {\scriptsize $\pm 0.18$} & 0.85 {\scriptsize $\pm 0.07$} & 0.63 {\scriptsize $\pm 0.06$} \\
    & Trust Score~\cite{NIPS2018_7798} & \underline{14.80} {\scriptsize $\pm 2.03$} & \underline{52.13} {\scriptsize $\pm 1.79$} & \underline{97.36} {\scriptsize $\pm 0.10$} & \underline{0.82} {\scriptsize $\pm 0.04$} & \underline{0.59} {\scriptsize $\pm 0.03$} \\
    & ConfidNet & \ubold{11.61} {\scriptsize $\pm 1.96$} & \ubold{59.72} {\scriptsize $\pm 1.90$} & \ubold{97.89} {\scriptsize $\pm 0.14$} & \ubold{0.70} {\scriptsize $\pm 0.05$} & \ubold{0.47} {\scriptsize $\pm 0.04$} \\
    \midrule
    \multirow{4}{*}{\shortstack[c]{\ubold{MNIST} \\ SmallConvNet}} & MCP~\cite{hendrycks17baseline} & 5.53 {\scriptsize $\pm 1.25$} & 36.08 {\scriptsize $\pm 3.60$} & 98.49 {\scriptsize $\pm 0.07$} & \underline{0.15} {\scriptsize $\pm 0.01$} & \underline{0.12} {\scriptsize $\pm 0.01$} \\
    & MC Dropout~\cite{Gal:2016:DBA:3045390.3045502} & \ubold{5.03} {\scriptsize $\pm 0.72$} & \underline{42.12} {\scriptsize $\pm 5.52$} & \underline{98.53} {\scriptsize $\pm 0.12$} & 0.16 {\scriptsize $\pm 0.01$} & \underline{0.12} {\scriptsize $\pm 0.01$} \\
    & Trust Score~\cite{NIPS2018_7798} & 9.60 {\scriptsize $\pm 2.69$} & 33.47 {\scriptsize $\pm 3.82$} & 98.20 {\scriptsize $\pm 0.23$} & 0.18 {\scriptsize $\pm 0.03$} & 0.15 {\scriptsize $\pm 0.02$} \\
    & ConfidNet & \underline{5.32} {\scriptsize $\pm 1.14$} & \ubold{45.45} {\scriptsize $\pm 3.75$} & \ubold{98.72} {\scriptsize $\pm 0.07$} & \ubold{0.13} {\scriptsize $\pm 0.02$} & \ubold{0.10} {\scriptsize $\pm 0.01$} \\
    \midrule
    \multirow{4}{*}{\shortstack[c]{\ubold{SVHN} \\ SmallConvNet}} & MCP~\cite{hendrycks17baseline} & \underline{32.17} {\scriptsize $\pm 0.91$} & \underline{46.20} {\scriptsize $\pm 0.50$} & \underline{92.93} {\scriptsize $\pm 0.13$} & \underline{5.58} {\scriptsize $\pm 0.14$} & \underline{4.50} {\scriptsize $\pm 0.09$} \\
    & MC Dropout~\cite{Gal:2016:DBA:3045390.3045502} & 33.54 {\scriptsize $\pm 1.06$} & 45.15 {\scriptsize $\pm 1.29$} & 92.84 {\scriptsize $\pm 0.08$} & 5.70 {\scriptsize $\pm 0.11$} & 4.61 {\scriptsize $\pm 0.09$} \\
    & Trust Score~\cite{NIPS2018_7798} & 34.01 {\scriptsize $\pm 1.11$} & 44.77 {\scriptsize $\pm 1.30$} & 92.65 {\scriptsize $\pm 0.29$} & 5.72 {\scriptsize $\pm 0.11$} & 4.64 {\scriptsize $\pm 0.12$} \\
    & ConfidNet & \ubold{29.90} {\scriptsize $\pm 0.76$} & \ubold{48.64} {\scriptsize $\pm 1.08$} & \ubold{93.15} {\scriptsize $\pm 0.15$} & \ubold{5.51} {\scriptsize $\pm 0.09$} & \ubold{4.43} {\scriptsize $\pm 0.08$} \\
    \midrule
    \multirow{4}{*}{\shortstack[c]{\ubold{CIFAR-10} \\ VGG16}} & MCP~\cite{hendrycks17baseline} & \underline{49.19} {\scriptsize $\pm 1.42$} & \underline{48.37} {\scriptsize $\pm 0.69$} & \underline{91.18} {\scriptsize $\pm 0.32$} & \underline{12.66} {\scriptsize $\pm 0.61$} & \underline{8.71} {\scriptsize $\pm 0.50$} \\
    & MC Dropout~\cite{Gal:2016:DBA:3045390.3045502} & 49.67 {\scriptsize $\pm 2.66$} & 48.08 {\scriptsize $\pm 0.99$} & 90.70 {\scriptsize $\pm 1.96$} & 13.31 {\scriptsize $\pm 2.63$} & 9.46 {\scriptsize $\pm 2.41$} \\
    & Trust Score~\cite{NIPS2018_7798} & 54.37 {\scriptsize $\pm 1.96$} & 41.80 {\scriptsize $\pm 1.97$} & 87.87 {\scriptsize $\pm 0.41$} & 17.97 {\scriptsize $\pm 0.45$} & 14.02 {\scriptsize $\pm 0.34$} \\
    & ConfidNet & \ubold{45.08} {\scriptsize $\pm 1.58$} & \ubold{53.72} {\scriptsize $\pm 0.55$} & \ubold{92.05} {\scriptsize $\pm 0.34$} & \ubold{11.78} {\scriptsize $\pm 0.58$} & \ubold{7.88} {\scriptsize $\pm 0.44$} \\
    \midrule
    \multirow{4}{*}{\shortstack[c]{\ubold{CIFAR-100} \\ VGG16}} & MCP~\cite{hendrycks17baseline} & 66.55 {\scriptsize $\pm 1.56$} & 71.30 {\scriptsize $\pm 0.41$} & 85.85 {\scriptsize $\pm 0.14$} & 113.23 {\scriptsize $\pm 2.98$} & 51.93 {\scriptsize $\pm 1.20$} \\
    & MC Dropout~\cite{Gal:2016:DBA:3045390.3045502} & \underline{63.25} {\scriptsize $\pm 0.66$} & \underline{71.88} {\scriptsize $\pm 0.72$} & \underline{86.71} {\scriptsize $\pm 0.30$} & \ubold{101.41} {\scriptsize $\pm 3.45$} & \ubold{46.45} {\scriptsize $\pm 1.91$} \\
    & Trust Score~\cite{NIPS2018_7798} & 71.90 {\scriptsize $\pm 0.93$} & 66.77 {\scriptsize $\pm 0.52$} & 84.41 {\scriptsize $\pm 0.15$} & 119.41 {\scriptsize $\pm 2.94$} & 58.10 {\scriptsize $\pm 1.09$} \\
    & ConfidNet & \ubold{62.70} {\scriptsize $\pm 1.04$} & \ubold{73.55} {\scriptsize $\pm 0.57$} & \ubold{87.17} {\scriptsize $\pm 0.21$} & \underline{108.46} {\scriptsize $\pm 2.62$} & \underline{47.15} {\scriptsize $\pm 0.95$} \\
    \bottomrule
  \end{tabular}
  \end{adjustbox}
\end{table*}

\subsection{Multi-scale ConfidNet architecture}
\label{sec:confidnet-multi}

In semantic segmentation, models consist of fully convolutional networks where hidden representations are 2D feature maps. This is in contrast with the architecture of classification models considered in Section \ref{sec:confidnet}. As a result, ConfidNet module must have a different design here: Instead of fully-connected layers, it is composed of $1\!\times\!1$ convolutional layers with the adequate number of channels.

In many segmentation datasets, the existence of objects at multiple scales may complicate confidence estimation.
As in recent works dealing with varying object sizes~\cite{ChenPK0Y16}, we further improve our confidence network $C$ by adding a multi-scale architecture based on spatial pyramid pooling. It consists of a computationally efficient scheme to re-sample a feature map at different scales, and then to aggregate the confidence maps.

From a feature map, we apply parallel atrous convolutional layers with $3\!\times\!3$ kernel size and different sampling rates, each of them followed by a series of 4 standard convolutional layers with $3\!\times\!3$ kernel size. In contrast with convolutional layers with large kernels, atrous convolution layers enlarge the field of view of filters and help to incorporate a larger context without increasing the number of parameters and the computation time. Resulting features are then summed before upsampling to the original image size of $H\times W$. We apply a final sigmoid activation to output a confidence map with values between 0 and 1.

The whole architecture of the confidence model $C$ is represented in the orange block of Figure~\ref{fig:confidence_training}, along with its training given a fixed segmentation model $F$ (blue block) with which it shares the encoder. Such as in previous section, fine-tuning the encoder within $C$ is also possible, although we did not explore the option in this semantic segmentation context due to the excessive memory overhead it implies.

\section{Experiments}
\label{sec:experiments}

We evaluate our approach on the two tasks presented in previous sections: Failure prediction in classification settings and semantic segmentation with domain adaptation. 

\subsection{Failure prediction}
\label{subsec:exp_confidnet}

In this section, we present comparative experiments against state-of-the-art confidence-estimation approaches and Bayesian methods on various datasets. Then, we study the effect of learning variants on our approach. 

\subsubsection{Experimental setup}

The experiments are conducted on image datasets of varying scale and complexity: MNIST \cite{lecun-mnisthandwrittendigit} and SVHN \cite{svhn-dataset} datasets provide small and relatively simple images of digits (10 classes). CIFAR-10 and CIFAR-100 \cite{Krizhevsky09} propose more complex object-recognition tasks on low resolution images. 

\begin{figure*}[t]
  \centering
  \includegraphics[width=0.7\linewidth]{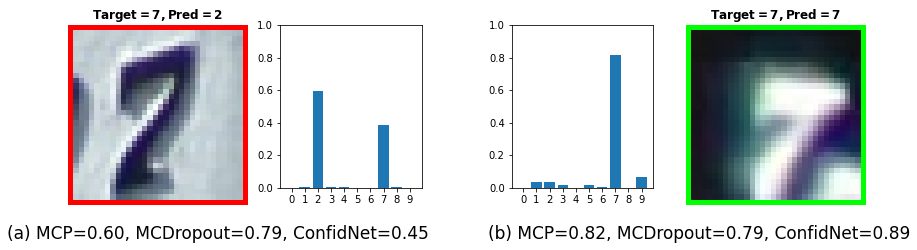}
  \caption{\textbf{Limitations of MC Dropout's confidence measure.}  
    Two test samples from SVHN dataset, which are respectively misclassified (left) and correctly classified (right) by a given model $F$, illustrate these limits. The entropies of the predicted class distributions (averaged over Monte Carlo dropout layers and displayed as histograms) are equally high, at around 0.79, resulting in equally low MC Dropout confidence estimates. In contrast, both MCP and TCP approximated by ConfidNet clearly differ as expected for the two examples. Yet, ConfidNet has the best behavior, being the lowest for the erroneous model's prediction and the highest for the correct one.}
  \label{visu-entropy}
\end{figure*}

\begin{table*}[t]
    \caption{\textbf{Impact of the choice of training data on the error-prediction performance of ConfidNet.} Comparison in AUPR between training on a model's train set or on a validation set.}
    \centering
    \label{validationset}
    \begin{adjustbox}{max width=\textwidth}
    \begin{tabular}{lcccccc}
        \toprule
        Variant & \ubold{MNIST} & \ubold{MNIST} & \ubold{SVHN} & \ubold{CIFAR-10} & \ubold{CIFAR-100} \\
        & MLP & SmallConvNet & SmallConvNet & VGG-16 & VGG-16 \\
        \midrule
        ConfidNet-\emph{train} & 59.72\% {\scriptsize $\pm 1.90$} & 45.45\% {\scriptsize $\pm 3.75$} & 48.64\% {\scriptsize $\pm 1.08$} & 53.72\% {\scriptsize $\pm 0.55$} & 73.55\% {\scriptsize $\pm 0.57$} \\
        ConfidNet-\emph{val}  & 38.22\% {\scriptsize $\pm 2.26$} & 31.90\% {\scriptsize $\pm 2.42$} & 43.15\% {\scriptsize $\pm 1.69$} & 53.01\% {\scriptsize $\pm 1.06$} & 73.82\% {\scriptsize $\pm 0.76$}  \\
        \bottomrule
    \end{tabular}
    \end{adjustbox}
\end{table*} 

The classification models range from small convolutional networks for MNIST and SVHN to the larger VGG-16 architecture for the CIFAR datasets. We also consider a multi-layer perceptron (MLP) with one hidden layer to investigate performances on small models. ConfidNet is attached to the penultimate layer of the convolutional neural network. Further details about datasets, architectures, training and metrics can be found in Appendix B.

We measure the quality of failure prediction following standard metrics used in the literature~\cite{hendrycks17baseline}: AUROC, the area under the receiver operating  characteristic; FPR\,@\,95\%\,TPR, the false-positive rate measured when the true-positive rate is 95\%; and AUPR, the area under the precision-recall curve, using here incorrect model's predictions as positive detection samples (see details in Appendix B.2). Among these metrics, AUPR is the most directly related to the failure detection task, and is thus the prevalent one in our assessment.

As an additional, indirect way to assess the quality of the predicted classifier's confidence, we also consider the selective classification problem that was discussed in Section \ref{sec:problem_formulation}. In this setup, the predictions by the classifier $F$ that get a predicted confidence below a defined threshold are rejected. Given a coverage rate (the fraction of examples that are not rejected), the performance of the classifier should improve. The impact of this selection, and hence of the underlying confidence-rate function, is measured in average with the area under the risk-coverage curve (AURC) and its normalized variant \emph{Excess}-AURC (E-AURC) \cite{geifman2018biasreduced}.

\subsubsection{Comparative results}

Along with our approach, we implemented competitive confidence and uncertainty estimation methods including MCP \cite{hendrycks17baseline}, Trust Score \cite{NIPS2018_7798}, and Monte-Carlo Dropout (MC Dropout) \cite{Gal:2016:DBA:3045390.3045502}. 

Comparative results are summarized in Table~\ref{comparative-results}. We observe that our approach outperforms the other methods in every setting, with a significant gap on small models/datasets. This confirms that TCP is an adequate confidence criterion for failure prediction and that our approach ConfidNet is able to learn it. Trust Score also delivers good results on small datasets/models such as MNIST. While ConfidNet still performs well on more complex datasets, Trust Score's performance drops, which might be explained by high-dimensionality issues with distances. Regarding selective classification results (AURC and E-AURC), we also provide risk-coverage curves in Appendix B.8 .

We also improve state-of-art performances from MC Dropout. While MC Dropout leverages ensembling based on dropout layers, taking as confidence measure the entropy on the average softmax distribution may not be always adequate. In Figure~\ref{visu-entropy}, we show side-by-side samples with similar distribution entropy. The left image is misclassified while the right one enjoys a correct prediction. In fact, the entropy is a permutation-invariant measure on discrete probability distributions: A correct 3-class prediction with score vector $[0.65, 0.34, 0.01]$ has the same entropy-based confidence as an incorrect one with probabilities $[0.34, 0.65, 0.01]$. In contrast, our approach can discriminate between an incorrect and a correct prediction, despite both having similarly-spread distributions. Note that, while fine-tuning ConfidNet doubles the overall computational complexity by using an auxiliary network with its own encoder, our approach is still better at failure prediction than a two-model ensemble (see. Appendix B.7).

\subsubsection{Effect of learning variants}
\label{subsec:learning_variants}

We analyse in Table~\ref{cloning-analysis} the effect of the encoder fine-tuning that is described in Section~\ref{subsec:confidnet-learning}. Learning only ConfidNet on top of the pre-trained encoder $E$ (that is, $\btheta = \bphi$), our confidence network already achieves significant improvements w.r.t. the baselines. With a subsequent fine-tuning of both modules (that is $\btheta = (\bw_{E'},\bphi))$, its performance is further boosted in every setting, by around 1-2\%. Note that using a vanilla fine-tuning without the deactivation of the dropout layers did not bring any improvement.

\begin{table}[ht]
\centering
  \caption{\textbf{Impact  of  the encoder fine-tuning  on  the  error-prediction  performance  of  ConfidNet}. Comparison in AUPR on two benchmarks with different backbones.}
  \label{cloning-analysis}
   \begin{adjustbox}{max width=\textwidth}
  \begin{tabular}{lcc}
    \toprule
    & \ubold{MNIST} & \ubold{CIFAR-100} \\
    & SmallConvNet & VGG-16\\
    \midrule
    Confidence training & 44.54\% & 71.30\% \\
    ~~+ Encoder fine-tuning & 45.45\% & 73.55\%\\
    \bottomrule
  \end{tabular}
  \end{adjustbox}
\end{table} 

\begin{table*}[t]
	\centering
	\caption{\textbf{Comparative performance on semantic segmentation with synth-to-real unsupervised domain adaptation.} Results in per-class IoU and class-averaged mIoU on GTA5\uda\,Cityscapes. All methods are based on a DeepLabv2 backbone.}
	\resizebox{\textwidth}{!}{%
		\begin{tabular}{l | c | c c c c c c c c c c c c c c c c c c c|c}
			\toprule
			\multicolumn{22}{c}{GTA5\uda\,Cityscapes}\\
			\toprule
			Method & \rotatebox{90}{Self-Train.} & \rotatebox{90}{road} & \rotatebox{90}{sidewalk} & \rotatebox{90}{building} & \rotatebox{90}{wall} & \rotatebox{90}{fence} & \rotatebox{90}{pole} & \rotatebox{90}{light} & \rotatebox{90}{sign} & \rotatebox{90}{veg} & \rotatebox{90}{terrain} & \rotatebox{90}{sky} & \rotatebox{90}{person} & \rotatebox{90}{rider} & \rotatebox{90}{car} & \rotatebox{90}{truck} & \rotatebox{90}{bus} & \rotatebox{90}{train} & \rotatebox{90}{mbike} & \rotatebox{90}{bike} & mIoU \\
			\midrule
			AdaptSegNet~\cite{Tsai_adaptseg_2018} & & 86.5 & 25.9 & 79.8 & 22.1 & 20.0 & 23.6 & 33.1 & 21.8 & 81.8 & 25.9 & 75.9 & 57.3 & 26.2 & 76.3 & 29.8 & 32.1 & 7.2 & \textbf{29.5} & 32.5 & 41.4 \\
			CyCADA~\cite{Hoffman_cycada2017} & & 86.7 & 35.6 & 80.1 & 19.8 & 17.5 & \textbf{38.0} & \textbf{39.9} & \textbf{41.5} & 82.7 & 27.9 & 73.6 & \textbf{64.9} & 19.0 & 65.0 & 12.0 & 28.6 & 4.5 & 31.1 & 42.0 & 42.7 \\
			DISE~\cite{chang2019all} & & 91.5 & 47.5 & 82.5 & 31.3 & 25.6 & 33.0 & 33.7 & 25.8 & 82.7 & 28.8 & 82.7 & 62.4 & 30.8 & 85.2 & 27.7 & 34.5 & 6.4 & 25.2 & 24.4 & 45.4 \\
			AdvEnt~\cite{vu2018advent} & & 89.4 & 33.1 & 81.0 & 26.6 & 26.8 & 27.2 & 33.5 & 24.7 & 83.9 & 36.7 & 78.8 & 58.7 & 30.5 & 84.8 & 38.5 & 44.5 & 1.7 & 31.6 & 32.4 & 45.5 \\	
			\midrule
		    CBST~\cite{zou2018unsupervised} & \checkmark & 91.8 & 53.5 & 80.5 & 32.7 & 21.0 & 34.0 & 28.9 & 20.4 & 83.9 & 34.2 & 80.9 & 53.1 & 24.0 & 82.7 & 30.3 & 35.9 & 16.0 & 25.9 & \textbf{42.8} & 45.9 \\
			MRKLD~\cite{Zou_2019_ICCV} & \checkmark & 91.0 & 55.4 & 80.0 & 33.7 & 21.4 & 37.3 & 32.9 & 24.5 & 85.0 & 34.1 & 80.8 & 57.7 & 24.6 & 84.1 & 27.8 & 30.1 & \textbf{26.9} & 26.0 & 42.3 & 47.1 \\
			BDL~\cite{Li_2019_CVPR} & \checkmark & 91.0 & 44.7 & 84.2 & 34.6 & \textbf{27.5} & 30.2 & 36.0 & 36.0 & 85.0 & \textbf{43.6} & 83.0 & 58.6 & \textbf{31.6} & 83.3 & 35.3 & 49.7 & 3.3 & 28.8 & 35.6 & 48.5 \\
			ESL~\cite{saporta2020esl} & \checkmark & 90.2 & 43.9 & 84.7 & 35.9 & 28.5 & 31.2 & 37.9 & 34.0 & 84.5 & 42.2 & 83.9 & 59.0 & 32.2 & 81.8 & 36.7 & 49.4 & 1.8 & 30.6 & 34.1 & 48.6 \\
			\rowcolor{Gray}
			ConDA & \checkmark & \textbf{93.5} & \textbf{56.9} & \textbf{85.3} & \textbf{38.6} & 26.1 & 34.3 & 36.9 & 29.9 & \textbf{85.3} & 40.6 & \textbf{88.3} & 58.1 & 30.3 & \textbf{85.8} & \textbf{39.8} & \textbf{51.0} & 0.0 & 28.9 & 37.8 & \textbf{49.9} \\
			\bottomrule
		\end{tabular}
	}
	\label{tab:conda-gta2cityscapes}
\end{table*}

Given the small number of erroneous-prediction samples that are available due to deep neural network over-fitting, we also experimented confidence training on a hold-out dataset. We report the results on all datasets in Table~\ref{validationset} for validation sets with 10\% of samples. We observe a general performance drop when using a validation set for training TCP confidence. The drop is especially pronounced for small datasets (MNIST), where models reach more than 97\% of train and validation accuracies. Consequently, with a high accuracy and a small validation set, we do not get a larger absolute number of errors using a hold-out set rather than the train set. One solution would be to increase the validation-set size, but this would damage the model's prediction performance. By contrast, we take care with our experiments to base our confidence estimation on models with levels of test predictive performance that are similar to those of the baselines. On CIFAR-100, the gap between train accuracy and validation accuracy is substantial (95.56\% vs. 65.96\%), which may explain the slight improvement for confidence estimation using a validation set (+0.17\%). We think that training ConfidNet on a validation set with models reporting low/medium test accuracies could improve the approach.

\begin{table}[ht]
  \caption{\textbf{Effect of the loss on the error-detection performance of ConfidNet.} Comparison in AUPR between proposed MSE loss and three other alternatives.}
  \centering
  \label{loss-analysis2}
  \begin{tabular}{crrrr}
    \toprule
    Dataset & MSE & BCE & Focal & Ranking \\
    \midrule
    \ubold{SVHN} & \ubold{50.72\%} & 50.00\% & 49.96\% & 48.11\% \\
    \ubold{CIFAR-10} & \ubold{49.94\%} & 47.95\% & 47.76\% & 44.04\% \\
    \bottomrule
  \end{tabular}
\end{table}

In Table~\ref{loss-analysis2}, we compare training ConfidNet with the MSE loss (\ref{eq:loss-conf}) to training with a binary-classification cross-entropy loss (BCE), a focal BCE loss and a batch-wise approximate ranking loss. Even though BCE specifically addresses the failure prediction task, it achieves lower performances on CIFAR-10 and SVHN datasets. Similarly, the focal loss and the ranking one yield results below TCP's performance in every tested benchmark. Our intuition is that TCP regularizes the training by providing finer-grain information about the quality of the classifier's predictions. This is especially important in the difficult learning configuration where only very few error samples are available due to the good performance of the classifier.

\subsection{Unsupervised domain adaptation in semantic segmentation}
\label{subsec:exp_conda}

\begin{figure*}[t]
\centering
\includegraphics[width=\linewidth]{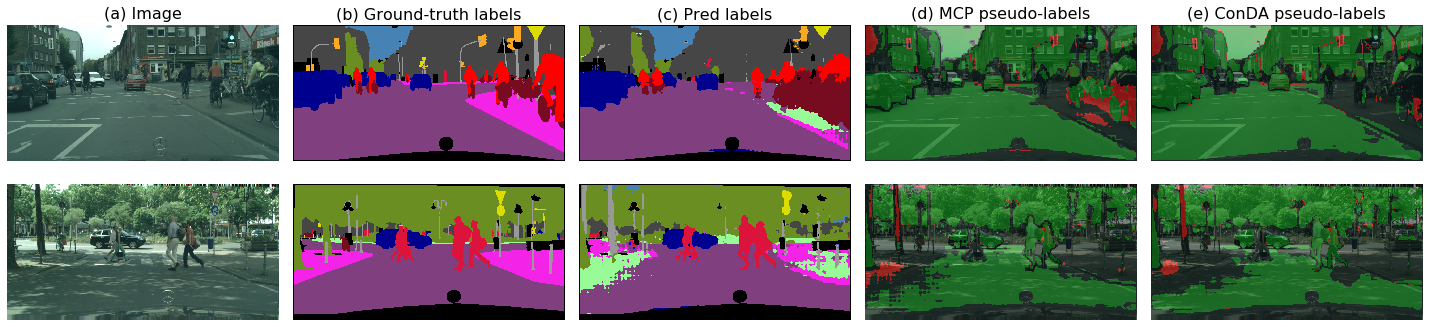}
\caption{\textbf{Qualitative results of pseudo-label selection for semantic-segmentation adaptation.} The three first columns present target-domain images of the GTA5\uda\,Cityscapes benchmark (a) along with their ground-truth segmentation maps (b) and the predicted maps before self-training (c). We compare pseudo-labels collected with MCP (d) and with ConDA (e). Green (resp. red) pixels are correct (resp. erroneous) predictions selected by the method and black pixels are discarded predictions. ConDA retains fewer errors while preserving approximately the same amount of correct predictions.}
\label{fig:qualitative_results}
\end{figure*}

In this section, we analyse on several semantic segmentation benchmarks the performance of ConDA, our approach to domain adaptation with confidence-based self-training. We report comparisons with state-of-the-art methods on each benchmark. We also analyse further the quality of ConDA's pseudo-labelling and demonstrate via an ablation study the importance of each of its components.

\subsubsection{Experimental setup}

As in many UDA works for semantic segmentation, we consider the specific task of adapting from synthetic to real data in urban scenes. We present in particular experiments in the common set-up, denoted GTA5\uda\,Cityscapes, where GTA5~\cite{richter-eecv2016} is the synthetic source dataset while the real-word target dataset is Cityscapes~\cite{cordts-cvpr2016}. We also validate our approach on two other benchmarks -- SYNTHIA\uda\,Cityscapes and SYNTHIA\uda\,Mapillary Vistas~\cite{neuhold-iccv2017} -- in Appendix C.3. The GTA5~\cite{richter-eecv2016} dataset is composed of 24,966 images extracted from the eponymous game, of dimension $1914 \times 1052$ and semantically annotated with 19 classes in common with Cityscapes~\cite{cordts-cvpr2016}. Cityscapes~\cite{cordts-cvpr2016} is a dataset of real street-level images. For domain adaptation, we use the training set as target dataset during training. It is composed of 2,975 images of dimension $2048 \times 1024$. All results are reported in terms of intersection over union (IoU) per class or mean IoU over all classes (mIoU); the higher this percentage, the better. 

We evaluate the proposed self-training method on AdvEnt~\cite{vu2018advent}, a state-of-the-art UDA approach. AdvEnt~\cite{vu2018advent} proposes an adversarial learning framework for domain adaptation: Instead of the softmax output predictions, AdvEnt aligns the entropy of the pixel-wise predictions. All the implementations are done with the PyTorch framework~\cite{NEURIPS2019_bdbca288}. The semantic segmentation models are initialized with DeepLabv2 backbones pretrained on ImageNet~\cite{krizhevsky2012imagenet}. Due to computational constraints, we only train the multi-scale ConfidNet without encoder fine-tuning. Further information about architectures and implementation details of training and metrics can be found in Appendix C.1.

\subsubsection{Comparison with state of the art}

The results of semantic segmentation on the Cityscapes validation set using GTA5 as source domain are available in Table~\ref{tab:conda-gta2cityscapes}. 
All the methods rely on DeepLabv2 as their segmentation backbone. We first notice that self-training-based methods from the literature are superior on this benchmark, with performance reaching up to $48.6\%$ mIoU with ESL~\cite{saporta2020esl}. ConDA outperforms all those methods by reaching $49.9\%$ mIoU. 

\subsubsection{Analysis}

\noindent \textbf{Ablation Study.} To study the effect of the adversarial training and of the multi-scale confidence architecture on the confidence model, we perform an ablation study on the GTA5\uda\,Cityscapes benchmark. The results on domain adaptation after re-training the segmentation network using collected pseudo-labels are reported in Table~\ref{tab:ablationstudy}. In this table, ``ConfidNet'' refers to the simple network architecture defined in Section~\ref{sec:confidnet} (adapted to segmentation by replacing the fully connected layers by $1\!\times\! 1$ convolutions of suitable width); ``Adv. ConfidNet'' denotes the same architecture but with the adversarial loss from Section \ref{sec:adv-loss} added to its learning scheme; ``Multi-scale ConfidNet'' stands for the architecture introduced in Section \ref{sec:confidnet-multi}; Finally, the full method, ``ConDA'' amounts to having both this architecture and the adversarial loss. We notice that adding the adversarial learning achieves significantly better performance, for both ConfidNet and multi-scale ConfidNet, with respectively $+1.4$ and $+0.8$ point increase. Multi-scale ConfidNet (resp. adv. multi-Scale ConfidNet) also improves performance up to $+0.9$ point (resp. $+0.3$) from their ConfidNet architecture counterpart. These results stress the importance of both components of the proposed confidence model.

\begin{table}[ht]
    \centering
	\resizebox{0.95\linewidth}{!}{%
	\begin{tabular}{l|cc|c}
		\toprule
		Model & Multi-Scale. & Adv & mIoU\\
		\midrule
		ConfidNet & & & 47.6 \\
		Multi-Scale ConfidNet  & \checkmark & & 48.5\\
		Adv. ConfidNet & & \checkmark & 49.0 \\
		ConDA (Adv. Multi-scale ConfidNet) & \checkmark & \checkmark & \ubold{49.9} \\
		\bottomrule
	\end{tabular}
	}
    \caption{\textbf{Ablation study on semantic segmentation with pseudo-labelling-based adaptation.} Full-fledged ConDA approach is compared on GTA5\uda\,Cityscapes to stripped-down variants (with/without multi-scale architecture in ConfidNet, with/without adversarial learning).}
    \label{tab:ablationstudy}
\end{table}

\begin{figure}[ht]
    \centering
    \includegraphics[width=0.9\linewidth,trim={0 0cm 0 0},clip]{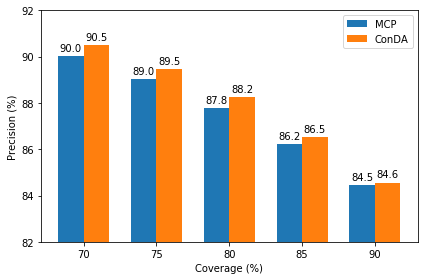}
    \vspace{-0.3cm}\caption{\textbf{Comparative quality of selected pseudo-labels}. Proportion of correct pseudo-labels (precision) for different coverages on GTA5\uda\,Cityscapes, for MCP and ConDA.}
    \label{fig:conda_analysis}
    \vspace{-0.5cm}
\end{figure}

\smallskip\noindent \textbf{Quality of pseudo-labels.}
Here we analyze the effectiveness of MCP and ConDA as confidence measures to select relevant pseudo-labels in the target domain. For a given fraction of retained pseudo-labels (coverage) on target-domain training images, we compare in Figure~\ref{fig:conda_analysis} the proportion of those labels that are correct (accuracy). ConDA outperforms MCP for all coverage levels, meaning it selects significantly fewer erroneous predictions for the next round of segmentation-model training. Along with the segmentation adaptation improvements presented earlier, these coverage results demonstrate that reducing the amount of noise in the pseudo-labels is key to learning a better segmentation adaptation model. Figure~\ref{fig:qualitative_results} presents qualitative results of those pseudo-labels methods. We find again that MCP and ConDA seem to select around the same amount of correct predictions in their pseudo-labels, but with ConDA picking out a lot fewer erroneous ones.

\section{Conclusion}
\label{sec:conclusion}

In this paper, we defined a new confidence criterion, TCP, which enjoys simple guarantees and empirical evidence of improving the confidence estimation for classifiers with a reject option. We proposed a specific method to learn this criterion with an auxiliary neural network built upon the encoder of the model that is monitored. Applied to failure prediction, this learning scheme consists in training the auxiliary network and then enabling the fine-tuning of its encoder (the one of the monitored classifier remains frozen). In each image classification experiment, we were able to improve the capacity of the model to distinguish correct from erroneous samples and to achieve better selective classification. Besides failure prediction, other applications can benefit from this improved confidence estimation. In particular, we showed that applied to self-training with pseudo-labels, our approach reaches state-of-the-art results on three synthetic-to-real unsupervised-domain-adaptation benchmarks (GTA5\uda Cityscapes, SYNTHIA\uda Cityscapes and SYNTHIA\uda Mapillary Vistas). To achieve these results, we equipped the auxiliary model with a multi-scale confidence architecture and supplemented the confident loss with an adversarial training scheme to enforces alignment between confidence maps in source and target domains. One limitation of this approach is the number of errors available in training. Further work includes exploring methods to artificially generate errors, such as aggressive data augmentation.


%




\ifCLASSOPTIONcaptionsoff
  \newpage
\fi



\bibliographystyle{IEEEtran}
\bibliography{IEEEabrv, references}

%


%

\begin{IEEEbiography}[{\includegraphics[width=1in,height=1.25in,clip,keepaspectratio]{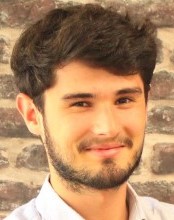}}]{Charles Corbi\`ere}
is a Ph.D. student in Deep Learning and Computer Vision  for Autonomous Driving at Conservatoire National des Arts et M\'etiers (CNAM, France) and valeo.ai research lab (France). He received an M.Sc. degree in Applied Mathematics and Statistic by Universit\'e Paris-Saclay (France) in 2017 and an M.Eng. degree in Computer Science from Ecole Centrale de Lille (France) in 2016. His research interests are uncertainty in deep learning and robustness.
\end{IEEEbiography}

\begin{IEEEbiography}[{\includegraphics[width=1in,height=1.25in,clip,keepaspectratio]{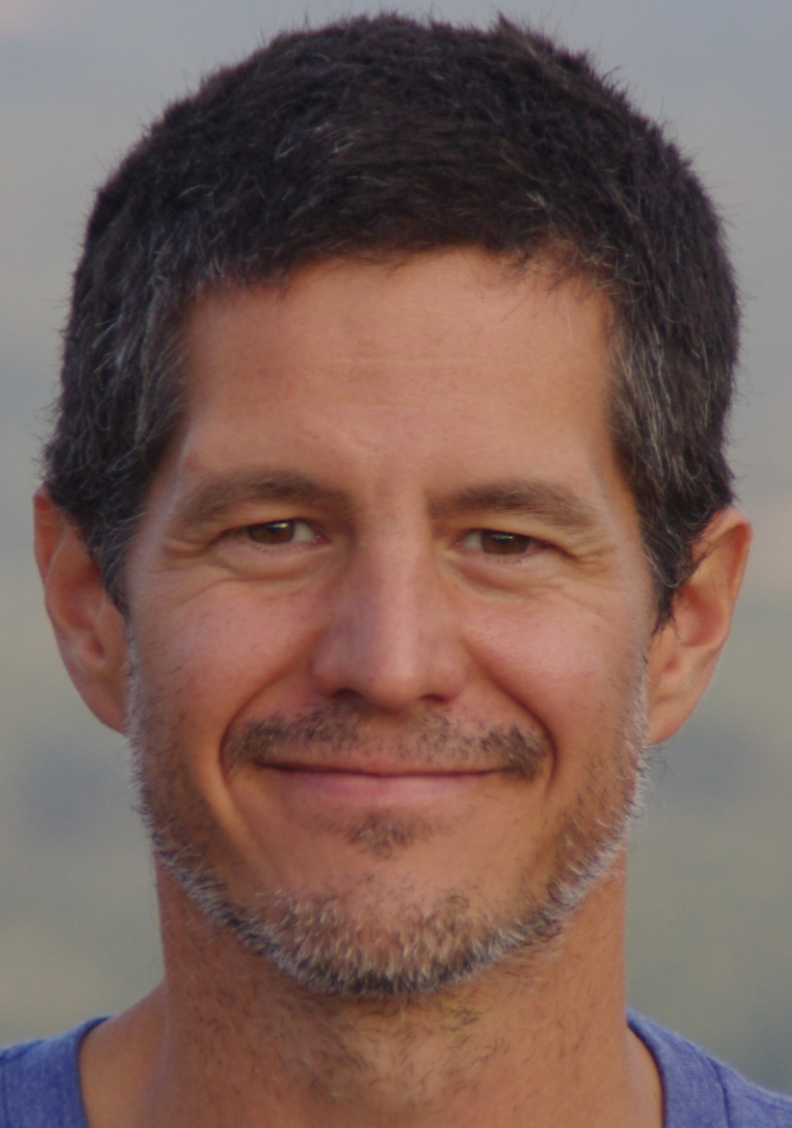}}]{Nicolas Thome} is a full professor at Conservatoire Nationnal des Arts et M\'etiers (Cnam Paris).
His research interests include machine learning and deep learning for understanding low-level signals, \eg. vision, time series, acoustics, etc. He also explores solutions for combining low-level and more higher-level data for multi-modal data processing. His current application domains are essentially targeted towards healthcare, autonomous driving and physics.  
He is involved in several French, European and international collaborative research projects on artificial intelligence and deep learning.
\end{IEEEbiography}

\begin{IEEEbiography}[{\includegraphics[width=1in,height=1.25in,clip,keepaspectratio]{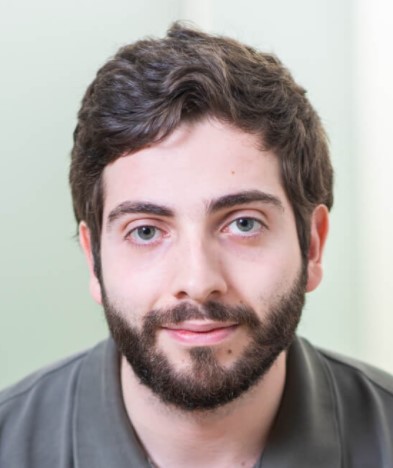}}]{Antoine Saporta}
is a Ph.D. student in Deep Learning and Computer Vision for Autonomous Driving at the Machine Learning and Deep Learning for  Information Access (MLIA) team of LIP6, Sorbonne Universit\'e (France) and Valeo.ai research lab. He is a graduate of \'Ecole Polytechnique (France) and has received a Master degree in Computer Science from Technische Universität München (Germany) in 2019. His research interests include domain adaptation and semantic segmentation.
\end{IEEEbiography}

\begin{IEEEbiography}[{\includegraphics[width=1in,height=1.25in,clip,keepaspectratio]{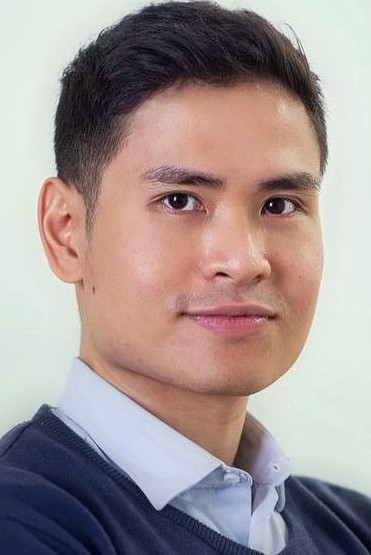}}]{Tuan-Hung Vu} is a research scientist at Valeo.ai. He received a PhD degree in Computer Science from \'Ecole Normale Sup\'erieure in 2018. His research interests include deep learning, object recognition, domain adaptation and more recently data augmentation. Tuan-Hung published and regularly served as reviewer at computer vision conferences and journals like CVPR, ICCV, ECCV and IJCV.
\end{IEEEbiography}

\begin{IEEEbiography}[{\includegraphics[width=1in,height=1.25in,clip,keepaspectratio]{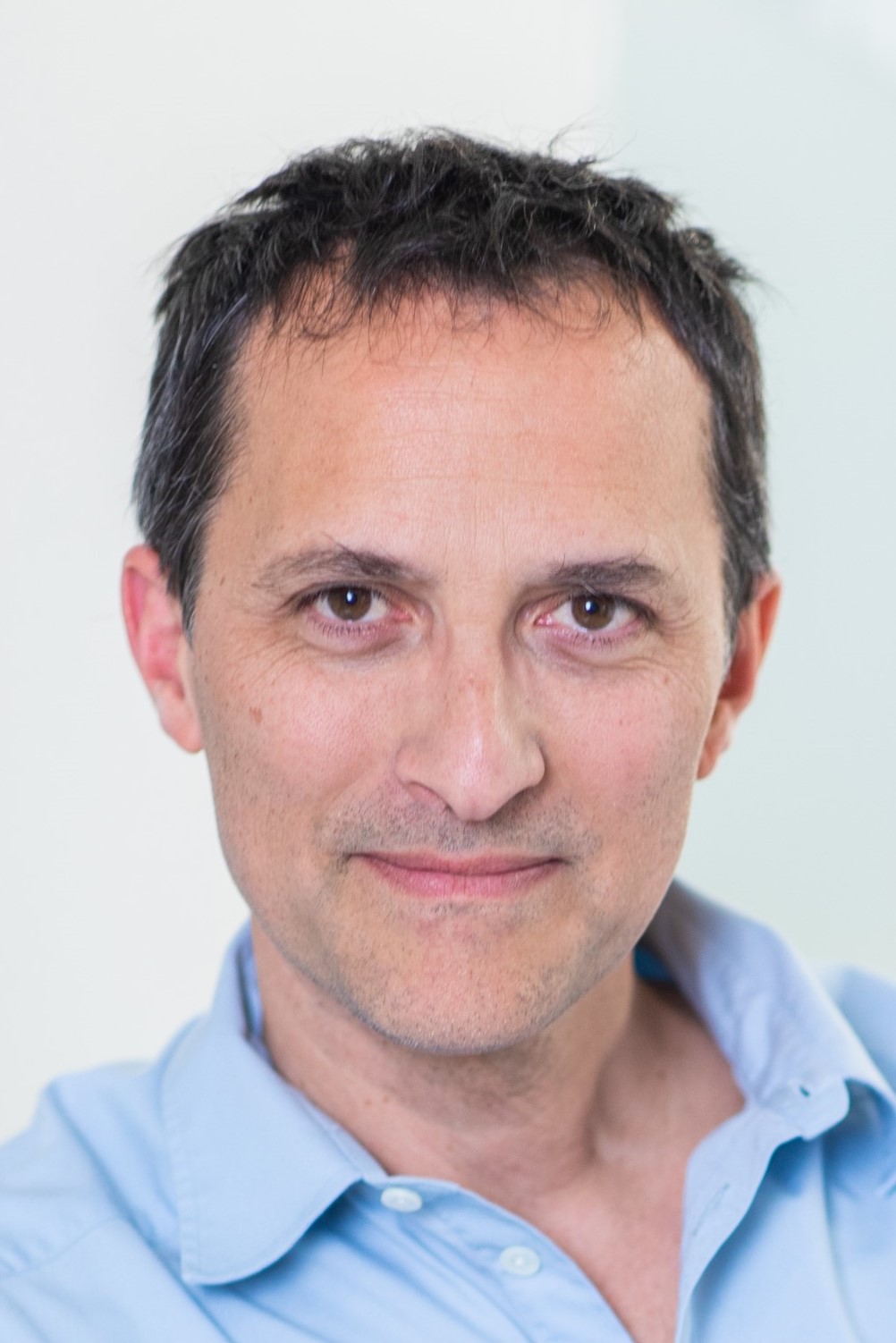}}]{Matthieu Cord} is full professor at Sorbonne University. He is also part-time principal scientist at Valeo.ai. His research expertise includes computer vision, machine learning and artificial intelligence. He is the author of more 150 publications on image classification, segmentation, deep learning, and multimodal vision and language understanding. He is an honorary member of the Institut Universitaire de France and served from 2015 to 2018 as an AI expert at CNRS and ANR (National Research Agency).
\end{IEEEbiography}

\begin{IEEEbiography}[{\includegraphics[width=1in,height=1.25in,clip,keepaspectratio]{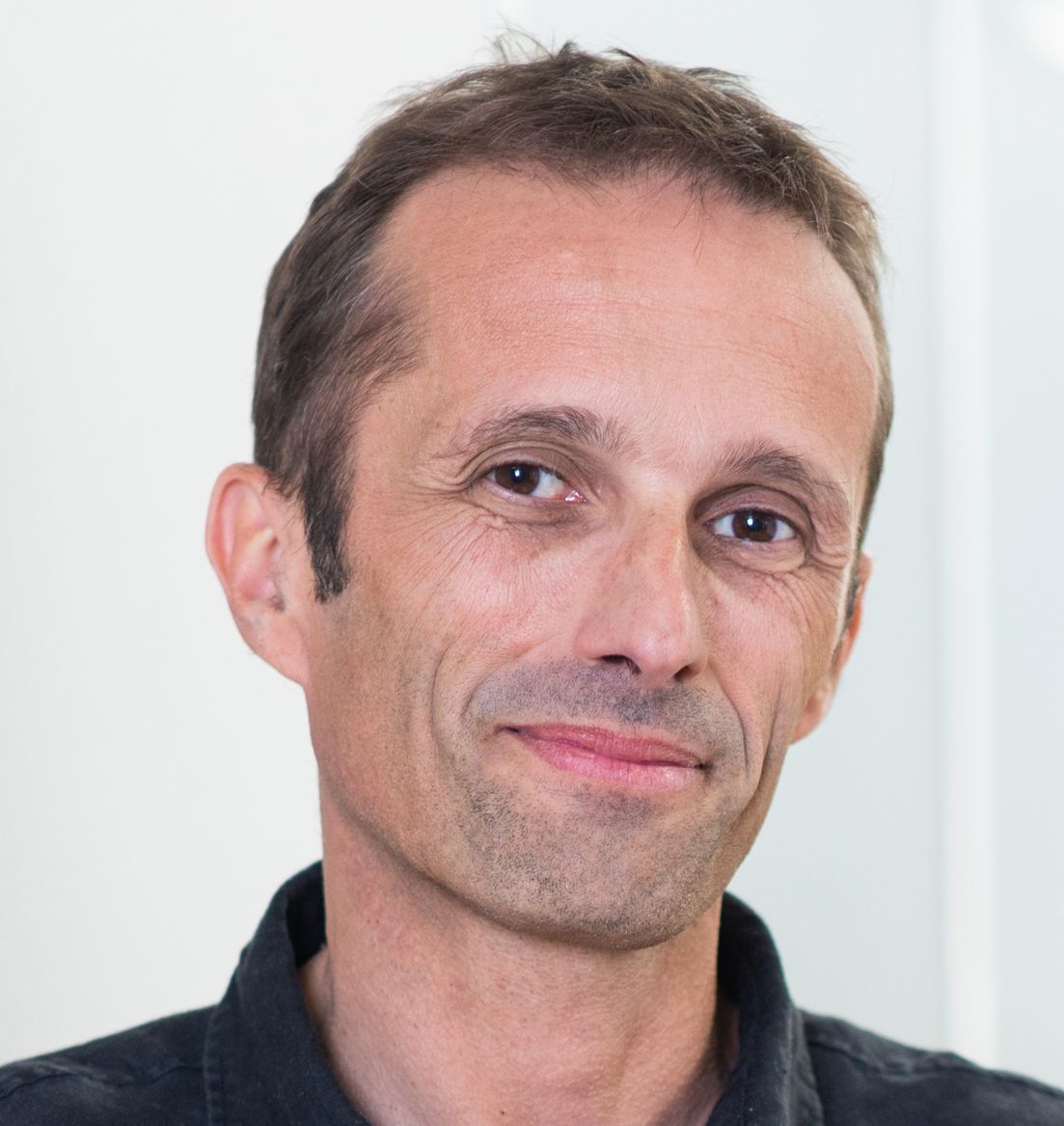}}]{Patrick P\'erez} is Scientific Director of Valeo.ai, a Valeo research lab on artificial intelligence for automotive applications. Before joining Valeo, Patrick P\'erez has been Distinguished Scientist at Technicolor (2009-2018), researcher at Inria (1993-2000, 2004-2009) and at Microsoft Research Cambridge (2000-2004). 
His research revolves around machine learning for scene understanding, data mining and visual editing.
\end{IEEEbiography}






\end{document}